\documentclass[12pt]{article}
\usepackage{arxiv}
\usepackage{amsmath}
\usepackage{graphicx,epstopdf}
\usepackage{amsmath}
\usepackage{times}
\usepackage{graphicx}
\usepackage{color}
\usepackage{multirow}
\usepackage{mathtools}
\usepackage{amssymb}
\usepackage{booktabs}       
\usepackage{amsfonts}       
\usepackage{nicefrac}       
\usepackage[authoryear]{natbib}	
\usepackage{tikz}
\usepackage{algorithm}
\usepackage{algpseudocode}
\usepackage{listings}
\usepackage{amsthm}
\newtheorem{claim}{Proposition}

\usepackage{tikz}
\usepackage{etoolbox} 
\usepackage{listofitems} 
\usepackage{tikz,ifthen}

\tikzset{>=latex} 
\colorlet{myred}{red!80!black}
\colorlet{myblue}{blue!80!black}
\colorlet{mygreen}{green!60!black}
\colorlet{mydarkred}{myred!40!black}
\colorlet{mydarkblue}{myblue!40!black}
\colorlet{mydarkgreen}{mygreen!40!black}
\tikzstyle{node}=[very thick,circle,draw=myblue,minimum size=22,inner sep=0.5,outer sep=0.6]
\tikzstyle{connect}=[<-,thick,mydarkblue,shorten >=1]
\tikzstyle{connect2}=[->,thick,mydarkblue,shorten >=1]
\tikzset{ 
  node 1/.style={node,mydarkgreen,draw=mygreen,fill=mygreen!25},
  node 2/.style={node,mydarkblue,draw=myblue,fill=myblue!20},
  node 3/.style={node,mydarkred,draw=myred,fill=myred!20},
}
\def\nstyle{int(\lay<\Nnodlen?min(2,\lay):3)} 

\title{Optimal Control with Natural Images: Efficient Reinforcement Learning using Overcomplete Sparse Codes}

\date{}

\author{Peter N.~Loxley\\
University of New England \\
Australia}

\begin{document}
\maketitle

\begin{abstract}

Optimal control and sequential decision making are widely used in many complex tasks. Optimal control over a sequence of natural images is a first step towards understanding the role of vision in control. Here, we formalize this problem as a reinforcement learning task, and derive general conditions under which an image includes enough information for optimal control. Reinforcement learning is shown to provide a computationally efficient method for finding optimal policies when natural images are encoded into ``efficient" image representations. This is demonstrated by introducing a new reinforcement learning benchmark that easily scales to large numbers of states and long horizons. In particular, by representing each image as an overcomplete sparse code, we are able to efficiently solve an optimal control task that is orders of magnitude larger than those tasks solvable using complete codes. Theoretical justification for this behaviour is provided. This work also demonstrates that deep learning is not necessary for efficient optimal control with natural images.

\end{abstract} 

 
\section{Introduction}\label{introduction} 

Many interesting and complex tasks can be described using natural image sequences. For optimal control with natural images, each image of the sequence corresponds to a particular state of an environment, and choosing an action or applying a control at the current state takes the environment to its next state, leading to the next image in the sequence. A natural question is then: given an image, what is the best control to apply? Although the answer will depend on the particular task at hand a general framework can be identified. As a simple example, consider moving through a rainforest and taking images of flowers in order to locate a particular species of plant. In this case, each image describes a state of the environment, and the state changes depending on the location of the next image. Each image also has a cost or reward associated with it: an image of the sought-after flower will receive a high reward, while an image without any flowers may receive a low (or even negative) reward. To perform this task well, the choice of control leading to the location of the next image does not necessarily give the best chance of obtaining an image of the sought-after flower at the next time period, rather, it should also lead to a good chance of obtaining such images at future time periods as you move through the rainforest. Planning over a sequence of time periods is the defining characteristic of sequential decision making and optimal control, and is a clear requirement of any intelligent system. 

Natural images form a high-dimensional dataset with a great deal of expressive power. For some tasks, images may include enough task-dependent information to determine an optimal policy so that no additional information is required for optimal control. Here, we will derive a general set of conditions for an image to be sufficient for implementing an optimal policy. Similar to other types of ecological data, it is well-known that the statistical properties of natural images distinguish them from artificially generated images \citep{Field1,Field2,ruderman1,ruderman2,simoncellio,hyvarinenbook,Hosseini}. We exploit these statistical properties in the following work. 

The aim of this work is to investigate the possibility of efficient optimal control over a sequence of natural images. An attractive idea is to make use of efficient neural representations of visual stimuli known as sparse codes. Statistical dependencies in natural images lead to redundancy, making it possible to obtain efficient representations using an appropriate encoding scheme such as sparse coding \citep{daugman3,daugman2,olshausenfield1,olshausenfield2}. A sparse code is ``efficient" in the sense of being a low bit-rate (low entropy) representation of a high-dimensional dataset. Here, we show an overcomplete sparse code is also \emph{computationally efficient} for finding optimal policies in reinforcement learning. 

The main idea is described by the neural network shown in Figure \ref{nn}. The first two layers of the neural network form a sparse autoencoder that generates an overcomplete sparse code by reconstructing an image input using an overcomplete basis of Gabor functions adapted to natural image statistics \citep{loxley1}. It is well-known that Gabor functions can be used to provide a ``sparse description" of natural images \citep{daugman3,daugman2}. A linear regression network then makes use of this overcomplete sparse code to approximate the cost-to-go required in reinforcement learning. The efficiency gain is twofold: an overcomplete sparse code is shown to increase both the learning rate and the storage capacity of a linear network used for storing cost-to-go values. In fact, both of these quantities can be improved by orders of magnitude. The underlying theoretical reason is that a sparse code favourably conditions the Hessian matrix of the least squares problem, while an overcomplete sparse code also helps to decorrelate image pixels so that columns of the associated design matrix are closer to being orthogonal.
\begin{figure}
\center
\begin{tikzpicture}[x=2cm,y=1cm]
  \readlist\Nnod{2,6,1} 
  \readlist\Nstr{d,m,} 
  \readlist\Cstr{I,\phi,r \boldsymbol{\cdot} \phi} 
  \def\yshift{0.55} 
  \foreachitem \N \in \Nnod{
    \def\lay{\Ncnt} 
    \pgfmathsetmacro\prev{int(\Ncnt-1)} 
    \foreach \i [evaluate={\c=int(\i==\N); \y=\N/2-\i-\c*\yshift;
                 \x=\lay; \n=\nstyle;
                 \index=(\i<\N?int(\i):"\Nstr[\n]");}] in {1,...,\N}{ 
      \node[node \n] (N\lay-\i) at (\x,\y) {$\strut\Cstr[\n]_{\index}$};  
      \ifnumcomp{\lay}{>}{1}{ 
        \foreach \j in {1,...,\Nnod[\prev]}{ 
          \draw[white,line width=1.2,shorten >=1] (N\prev-\j) -- (N\lay-\i);
          \ifthenelse{\N > 1}{
          \draw[connect] (N\prev-\j) -- (N\lay-\i);}
          {\draw[connect2] (N\prev-\j) -- (N\lay-\i);}    
        }
      }   
    }
    \ifthenelse{\N > 1}{
    \path (N\lay-\N) --++ (0,1+\yshift) node[midway,scale=1.6] {$\vdots$}; 
    }{}
  }
\node[above=3,align=center,mydarkgreen] at (N1-1.90) {Image\\ Input\\};
\node[above=2,align=center,mydarkblue] at (N2-1.90) {Overcomplete\\Sparse\\ Code\\};
\node[above=3,align=center,mydarkred] at (N\Nnodlen-1.90) {Cost-To-Go\\\vspace{15pt}};
\end{tikzpicture}
\caption{Neural network for reinforcement learning with natural images. The first two layers form a sparse autoencoder that generates an overcomplete sparse code $\phi$ by reconstructing the image input $I$ using an overcomplete basis of Gabor functions adapted to natural image statistics. The output $r\boldsymbol{\cdot}\phi$ then approximates the cost-to-go $\beta$ using weights $r$. The network storage capacity has increased from $d$ to (close to) $m$ by using an overcomplete sparse code. }
\label{nn}
\end{figure}

In this work, we formalize optimal control with natural images as a reinforcement learning task, and derive the general conditions under which an image includes sufficient information for optimal control. To prove (empirically) that an overcomplete sparse code gives the most efficient image representation compared with alternative approaches, we develop a new reinforcement learning benchmark. This benchmark allows the performance and efficiency of different image representations to be compared on a scaleable optimal control task. As no well-established benchmark for optimal control with natural images currently exists, this new benchmark fills a void. The benchmark is scalable so the number of states and length of horizon of the optimal control task can be easily increased; and the benchmark design ensures both optimal and suboptimal solutions coexist so that optimality can be clearly demonstrated. 

\begin{figure}[] 
\centering
\includegraphics[scale=0.15]{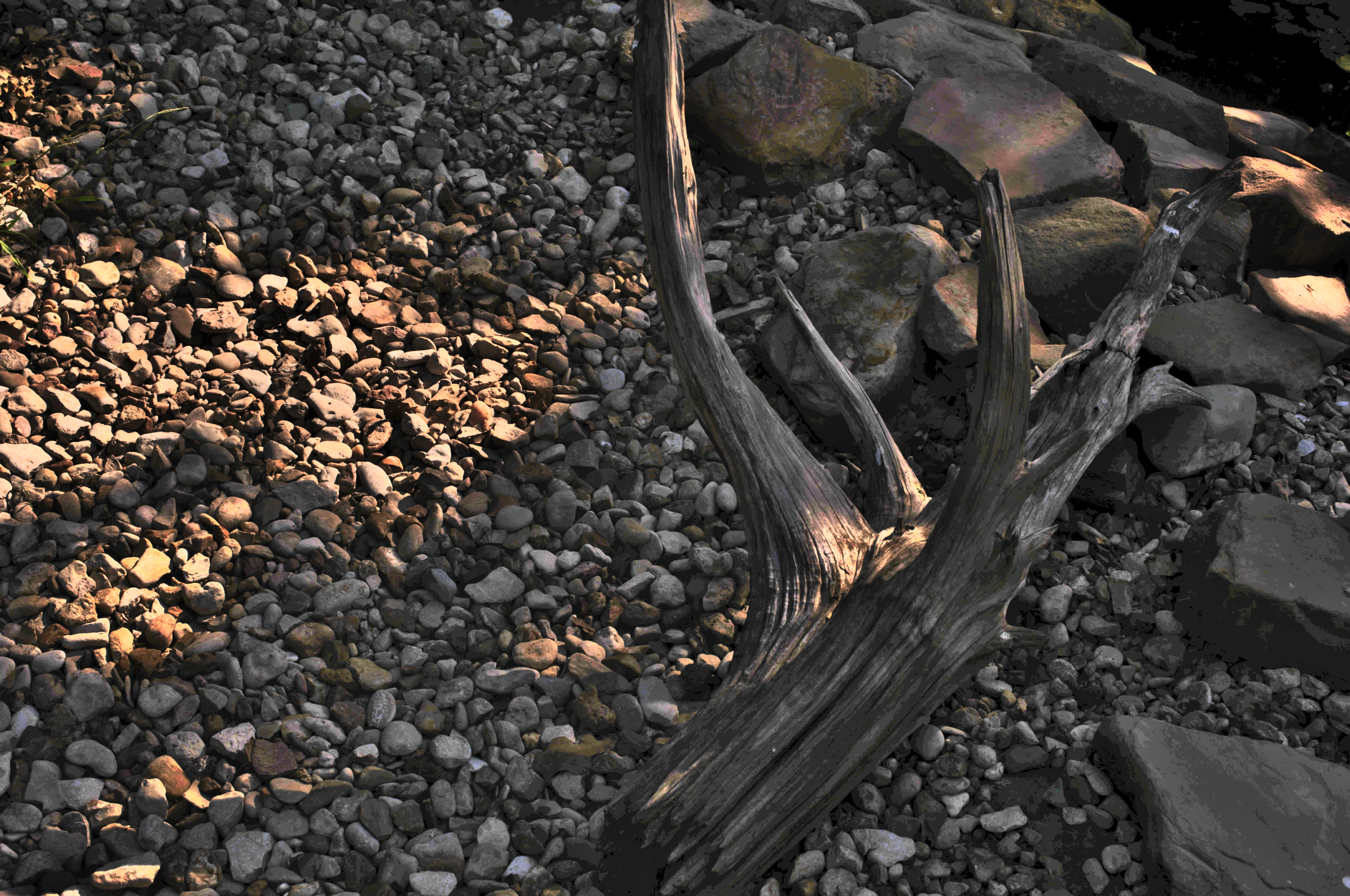}
\caption{A typical image used in the reinforcement learning benchmark (taken from the Geisler and Perry database).}
\label{nat_image}
\end{figure}
Previous work on reinforcement learning is heavily concentrated on synthetic datasets of limited complexity (although the tasks themselves may have very high complexity). Researchers have previously suggested that more work should be done to extend the paradigm of reinforcement learning to real-world data such as natural images (as shown in Figure~\ref{nat_image}, for example) and natural video \citep{zhang2018natural}. The few available examples of these efforts include using image representations to integrate unsupervised training of deep neural networks into reinforcement learning; specifically for visual navigation \citep{lange}, and simulated autonomous driving tasks \citep{zhang2021learning}. A little more work exists on the role of sparse codes and sparse representations in reinforcement learning, such as work to obtain better policies with neural networks on certain standard classic control tasks \citep{Martha1,Martha2,Rafati}. This line of work dates back to early research on coarse coding in reinforcement learning. However, none of these works considered sparse codes of images or other high-dimensional datasets of high complexity. The closest previous work appears in \cite{loxley2}, where sparse representations of images were applied to find optimal policies for a specific control task. Unfortunately, this work was limited by the size of the task; which required a hand-annotated video dataset, and the type of task; which was restricted to deterministic controllable dynamics rather than a general Markov decision process. These limitations did not allow results to be extended to large numbers of states and long horizons, or to non-deterministic control problems. In this work, a more general approach to optimal control is taken, and the development of a new reinforcement learning benchmark now allows results to be easily scaled up.

The structure of this paper is as follows. In Sec.~\ref{theory}, reinforcement learning methodology is briefly summarized, and the problem of optimal control over a sequence of natural images is formalized. Following this, a new reinforcement learning benchmark is developed in Sec.~\ref{generalbenchmark}. In Sec.~\ref{results}, results for the reinforcement learning benchmark are presented, and alternative image representations are compared. In Sec.~\ref{conc}, a detailed discussion and conclusion are given, and an analysis of algorithm time and memory efficiency is provided.

\section{Reinforcement Learning for Natural Image Tasks}\label{theory}
We consider a general optimal control task involving a discrete time dynamical system with a finite number of states and controls. During any time period $k$, the environment is described by a state $i$ taken from a state space $S$, and a control $u$ is selected from a set of available controls $U_k(i)$ (we adopt notation from \cite{bert2}). The state of the environment changes when a controller or agent chooses control $u$ from state $i$, moving the environment to state $j$ according to the transition probabilities $p_{ij}(u)$ of a controllable Markov chain. The cost of this transition is given by $g_k(i,u,j)$. An optimal policy $\mu_k^*(i)$ gives the optimal choice of control $\mu_k^*(i)=u^*$ at each state $i$ and time period $k$ that minimizes the expected value of the sum of costs over all time periods (i.e., over the horizon) of the control problem. A finite-horizon optimal control task can be solved exactly using the \emph{dynamic programming algorithm} \citep{bellman}, leading to the optimal cost-to-go $J_{k}^*$ (this algorithm is called \emph{value iteration} in the case of infinite horizon tasks, and converges to $J^*$ for discounted problems with stationary policies \citep{bert2}). An optimal policy $\mu_k^*(i)$ is then given by the minimizing control $u^*$ in the expression:
\begin{equation}
\underset{u\in U_k(i)}{\operatorname{min}}\ \sum_j p_{ij}(u)\left ( g_k(i,u,j)+J_{k+1}^*(j)\right ),\label{exact}
\end{equation}
for each state $i$, and each time period $k$. 

Optimal control with natural image sequences requires a slight modification of this general framework. The key consideration is now what information to include in a state to allow for optimal decision making. Any state that includes all necessary information for optimal control is called a \emph{sufficient statistic} \citep{bert2}. The main assumption we make here is that an image forms a sufficient statistic. In this case, each state $i\in S$ in an optimal control task is now given by an image (or image representation) $\phi(i)\in \bar{S}$; where $\bar{S}\subset \mathbb{R}^p$ so that each image is represented as a $p$-dimensional vector. An image is a sufficient statistic if $U_k(i),p_{ij}(u)$, and $g_k(i,u,j)$ in (\ref{exact}) depend on $i$ or $j$ only through $\phi(i)$ or $\phi(j)$, respectively. In other words, the index $i$ now only denotes a particular image in the modified framework and is not part of the information required for optimal control. For example, a striking consequence of $\phi(i)$ as a sufficient statistic is the form the transition probabilities now take:
\begin{equation}
\bar{p}(\phi(j)|\phi(i),u),
\end{equation}
so that given image $\phi(i)$, selecting control $u$ leads to image $\phi(j)$ with probability $\bar{p}(\phi(j)|\phi(i),u)$. Therefore, a controllable Markov chain now becomes an ``image generator" when it is used to generate new samples according to $\phi(j)\sim\bar{p}(.|\phi(i),u)$. We present a simple ``image generator" in the next section. The main consequence of $\phi(i)$ as a sufficient statistic is that the optimal policy is now written as 
\begin{equation}
\bar{\mu}_k^*(\phi(i)),\label{suff}
\end{equation}
and therefore optimal control depends only on the information contained in image $\phi(i)$.

We now present an algorithm for finding an optimal policy in this modified framework. Given an image $\phi(i)$ we require a function $\bar{J}_k(\phi(i))$, analogous to $J_k^*(i)$, that determines its cost-to-go. A simple choice of function is a linear network given by the scalar product of a weight vector $r_k\in\mathbb{R}^p$ with an image vector $\phi(i)$:
\begin{equation} 
\bar{J}_k(\phi(i),r_k)=r^\top_k \phi(i),\label{linear}
\end{equation}
where $\top$ is the vector transpose. To approximate the optimal cost-to-go $J_{k}^*$ requires adjusting the network weights using a training rule and a dataset. We apply \emph{fitted value iteration} to this task. To start the iteration, we go to the end of the task at $k=N-1$, and make use of the known terminal cost $\bar{J}_N(\phi(i),r_N^*)=g_N(i)$. A pair of expressions are then iterated backwards from $k=N-1$ to $k=0$ using $\bar{J}_N$ to start the iteration. The first expression is:
\begin{equation}
\beta_k^i = \underset{u\in \bar{U}_k(\phi(i))}{\operatorname{min}}\ \sum_{j} \bar{p}(\phi(j)|\phi(i),u)\left \{ \bar{g}_k(\phi(i),u,\phi(j))+\bar{J}_{k+1}(\phi(j),r_{k+1}^*)\right \}
, \label{iter1}
\end{equation}
and yields the cost-to-go $\beta_k^i$ of image $\phi(i)$ at time period $k$, by adding the cost of the current transition: $\bar{g}_k(\phi(i),u,\phi(j))$, to the cost-to-go of the next transition: $\bar{J}_{k+1}(\phi(j),r_{k+1}^*)$ to image $\phi(j)$ at time period $k+1$. The cost-to-go therefore accumulates the sum of costs over the horizon of the task. The second expression is:
\begin{align}
&\text{minimize}\ \ \ \sum_{i} (\bar{J}_{k}(\phi(i),r_k) - \beta_k^i)^2,\label{iter2}\\
&\text{subject to}\ \ \ r_k\in\mathbb{R}^p,\nonumber
\end{align}
and fits the function $\bar{J}_{k}(\phi(i),r_k)$ to the cost-to-go $\beta_k^i$ given a set of data points $(\phi(i),\beta_k^i)$ from (\ref{iter1}), by adjusting the weights $r_k$ to minimize the sum of squares. These iterations yield the cost-to-goes $\bar{J}_{N-1},...,\bar{J}_{0}$; as well as the optimal policy $\{\bar{\mu}_0^*,...,\bar{\mu}_{N-1}^*\}$ for the function approximation $\bar{J}_{k}(\phi(i),r_k)$.

\section{A Scalable Benchmark for Efficient Image Representations}\label{generalbenchmark}

Natural images used for optimal control in (\ref{linear})--(\ref{iter2}) are  computationally inefficient when using raw image pixels, as will be demonstrated shortly. It turns out to be much better to use some type of image representation. We now present a simple and robust benchmark for the purpose of comparing alternative image representations for optimal control tasks. This benchmark is designed to satisfy three important requirements: 1) the number of states should scale to arbitrarily large numbers, 2) the dynamics and cost structure should scale to arbitrarily long horizons and be easy to implement, and 3) greedy policies should be suboptimal. 

The number of states of the benchmark can scale to arbitrarily large numbers by working with \emph{image patches}. An image patch is an arbitrarily sized image region extracted from a larger parent image. Discretizing a parent image into a regular grid pattern yields a set of image patches $\bar{S}$, with each image patch $\phi(i)\in\bar{S}$ labeled by index $i$. Given a parent image of fixed size, the number of image patches can be increased simply by decreasing the size of each image patch. 

The following proposition will now be used to help construct the benchmark.
\begin{claim}
A state of the benchmark can be described by $i$ or $\phi(i)$, and either state is a sufficient statistic.
\end{claim}
\renewcommand\qedsymbol{$\blacksquare$}
\begin{proof}
Given any image patch $\phi(i)\in \bar{S}$, it is possible to determine its index $i$ by running through all image patches to find the closest match:
$$\hat{i}=\underset{m}{\operatorname{argmin}}\lVert \phi(m) - \phi(i)\rVert_2^{2},$$
provided each image patch is unique. This index also corresponds to the state $i\in S$. Alternatively, given state $i$, it is possible to determine the image patch $\phi(i)$ by looking up the $i$th element of $\bar{S}=\{\phi(1),...,\phi(n)\}$. The same is true for both $j$ and $\phi(j)$. Therefore, given $p_{ij}(u)$, $g_k(i,u,j)$, and $U_k(i)$; it is possible to determine $\bar{p}(\phi(j)|\phi(i),u)=p_{ij}(u)$, $\bar{g}_k(\phi(i),u,\phi(j))=g_k(i,u,j)$, and $\bar{U}_k(\phi(i))=U_k(i)$. Alternatively, given $\bar{p}(\phi(j)|\phi(i),u)$, $\bar{g}_k(\phi(i),u,\phi(j))$, and $\bar{U}_k(\phi(i))$; it is possible to determine $p_{ij}(u)=\bar{p}(\phi(j)|\phi(i),u)$, $g_k(i,u,j) = \bar{g}_k(\phi(i),u,\phi(j))$, and $U_k(i)=\bar{U}_k(\phi(i))$.
\end{proof}

Proposition 1 allows us to complete the benchmark as follows. In Appendix A1, a benchmark target tracking task is developed that satisfies all three of the important requirements listed above. This task provides us with $p_{ij}(u)$, $g_k(i,u,j)$, and $U_k(i)$. Proposition 1 can then be applied to obtain $\bar{p}(\phi(j)|\phi(i),u)$, $\bar{g}_k(\phi(i),u,\phi(j))$, and $\bar{U}_k(\phi(i))$. After substituting these quantities into (\ref{iter1}), we get:
\begin{equation}
\beta_k^i=\lVert a_1\rVert_2^2 + \underset{u\in U}{\operatorname{min}}\ \sum_{b_2} p(b_2|b_1)\bar{J}_{k+1}(\phi(a_1+u-b_2,b_2),r_{k+1}^*),\label{bmt}
\end{equation}
where 
\begin{equation}
i=(a_1,b_1)\in D\times T,
\end{equation}
and
\begin{equation}
j=(a_2,b_2).
\end{equation}
The sets $T$ and $U$ are specified by $T=\{(0,0),(1,1),(0,1)\}$ and $U=\{(0,0),(1,0),(0,1)\}$, while the set $D\subset\mathbb{Z}^2$ depends on the number of states (image patches) chosen for the benchmark. The transition probabilities $p$ correspond to the Markov chain described in Appendix A1. Once $D$ has been specified, an optimal policy for the benchmark is found by applying fitted value iteration using expressions (\ref{linear}), (\ref{iter2}), and (\ref{bmt}).

The design requirements of the benchmark are seen to be satisfied as follows. The number of states scales with the (arbitrary) size of the set $D$. The length of the horizon (number of time periods) scales with $N$. Finally, a greedy policy is distinct from the optimal policy due to the specific form chosen for $p$ in (\ref{bmt}). Most importantly, the benchmark can be used to compare the performance of any image representation $\phi(i)$, or any neural network architecture $\bar{J}_{k}(\phi(i),r_k)$ for optimal control.

\section{Results}\label{results}

All code for generating the figures in Section 4 is available at \cite{Loxley4}.

\subsection{Efficient Image Representations for Reinforcement Learning}

Image representations were found by transforming image patches from the benchmark into \emph{overcomplete sparse codes} and \emph{whitened complete codes}. A set of 48 natural images was taken from the database ``Natural Scene Statistics in Vision Science" (Set 5) by \cite{geisler}. A typical image from this dataset is shown in Figure \ref{nat_image}. Each image was cropped to $2844\times 2844$ pixels, then converted to grayscale and double precision using the Matlab functions \texttt{rgb2gray} and \texttt{im2double}, before being discretized into square image patches of side-length $a$ (measured in pixels). The number of image patches in a single image was therefore $\lfloor 2844/a\rfloor^2$; where the function $\lfloor .\rfloor$ rounds down to the nearest integer. 

Overcomplete sparse codes were constructed using the sparse autoencoder shown in Figure \ref{nn}. Starting with an image given by a vector of length $d$, the aim of an overcomplete sparse code is to generate an image representation given by a vector of length $p=m>>d$. This is done in the first stage of the autoencoder by solving a linear least-squares problem: minimize $\lVert G\phi - I\rVert_2^{2}$ with respect to $\phi\in\mathbb{R}^m$, given $I\in\mathbb{R}^d$ and a matrix $G\in\mathbb{R}^{d\times m}$. Here, $G$ is formed from an overcomplete basis of Gabor functions adapted to natural image statistics as detailed in Appendix A2. The two-dimensional Gabor function was originally used to model simple-cell receptive field profiles in the primary visual cortex \citep{hubel,jones,daugman1,daugman2}. The second stage of the neural network in Figure \ref{nn} gives an approximation (\ref{linear}) of the cost-to-go by solving another linear least-squares problem: minimize $\lVert r\boldsymbol{\cdot} \phi - \beta \rVert_2^{2}$ with respect to $r\in\mathbb{R}^m$, given $\phi$ from the first stage and $\beta$ from (\ref{bmt}). This linear least-squares problem is given in (\ref{iter2}).

The size of a sparse code depends on the number of Gabor functions in the overcomplete basis used to reconstruct an image input. To generate a $\times 64$ overcomplete sparse code, the number of Gabor functions required is $64\times a^2$ for an input image patch of $a^2$ pixels. In order to manage the storage capacity of our network we enforce the inequality $m>n$; where $m$ is the size of an overcomplete sparse code, and $n$ is the number of images (i.e., the number of cost-to-goes stored in the network). For this reason, we choose $a$ to maximize the number of image patches available in a $2844\times 2844$ image, whilst ensuring the size of a $\times 64$ overcomplete sparse code is larger than this number; i.e., $64\times a^2>\lfloor 2844/a\rfloor^2$. The unique solution to this problem is $a=19$, leading to 22201 image patches of 361 pixels each. 

A more traditional method of pixel decorrelation (whitening) is used to construct complete codes: where $p=m=d$. This method is carried out by setting the mean pixel values (taken over all image patches) to zero, and diagonalizing the corresponding covariance matrix. This method does not generalize to overcomplete codes. The second stage of the neural network in Figure \ref{nn} is then applied to approximate the cost-to-go.

\subsection{Benchmark Performance with Number of States}

\begin{figure}[] 
\centering
\includegraphics[scale=0.65,clip=true,bb=100 290 500 600]{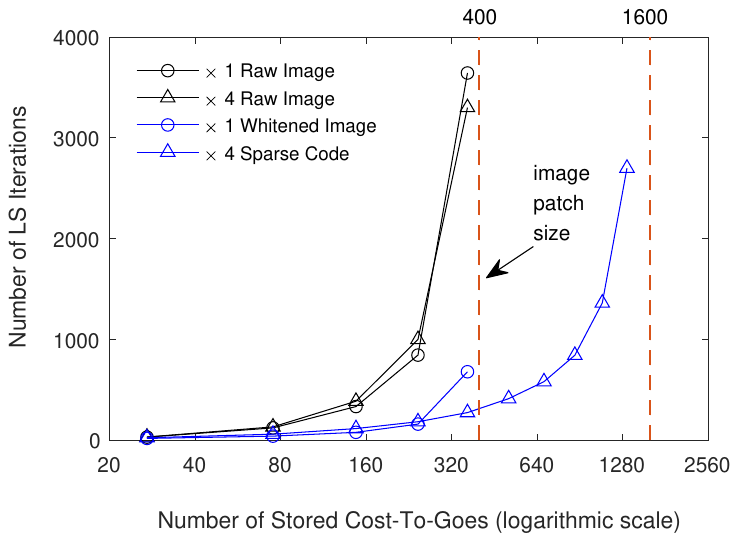}
\caption{Number of least squares iterations to store a number of cost-to-go values in a linear network using different image patch representations (see legend). The least squares objective is found within a tolerance of $1\times 10^{-6}$, and the number of iterations are averaged over 48 independent trials. Each image patch is 400 pixels in this figure.}
\label{result1}
\end{figure}

\begin{figure}[] 
\centering
\includegraphics[scale=0.65,clip=true,bb=50 290 500 600]{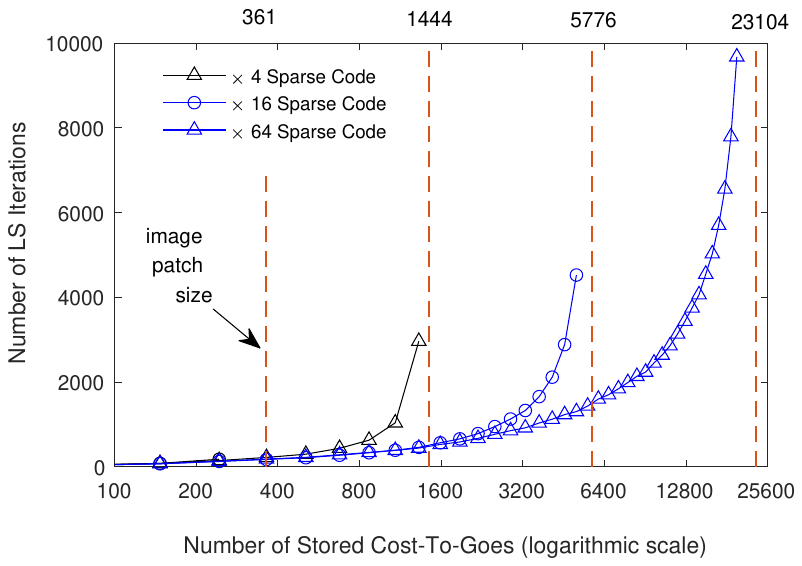}
\caption{Number of least squares iterations to store a number of cost-to-go values in a linear network using different image patch representations (see legend). The least squares objective is found within a tolerance of $1\times 10^{-6}$, and the number of iterations are averaged over 48 independent trials. Each image patch is 361 pixels.}
\label{result2}
\end{figure}

Training a neural network on efficient image representations takes fewer computational resources than training it directly on images. This is shown in Figs.~\ref{result1} and \ref{result2}, where the number of least squares iterations required to store a given number of cost-to-go values in a linear network is shown for network inputs given by raw image patches, and various representations of image patches. Each state of the benchmark is a unique image patch associated with a cost-to-go value. By adjusting the network weights during training, this association is stored in the network to determine an optimal policy. In a linear network, the number of network weights determines its storage capacity: the maximum number of cost-to-go values that can be stored. In turn, the number of weights is equal to the number of pixels in the input to the network; since each pixel input is multiplied by a weight and then summed together to yield an output. Therefore, the storage capacity of a network can be increased only by increasing the number of pixel inputs and associated network weights. One way to do this for a fixed-sized image patch is to generate an overcomplete representation, and use this as input to the network instead -- shown as the first layer in Figure 1. This approach will be used to solve the benchmark tracking task as the number of states and stored cost-to-go values increases. Network training is carried out using the Matlab \texttt{lsqr} function running on an Nvidia GPU, with the number of least-squares iterations until convergence (within $1\times 10^{-6}$ of the least-squares objective), averaged over 48 independent trials, shown along the vertical axis of Figs.~\ref{result1} and \ref{result2}. 

In Figure~\ref{result1}, the size of each image patch was chosen to be 400 pixels instead of 361 pixels for visualization purposes when the number of stored cost-to-goes is relatively small. When the network input is either a raw image patch or a complete representation of an image patch, a linear network will have exactly 400 inputs and 400 weight parameters, allowing it to store a maximum of 400 cost-to-go values. The curves for a $\times 1$ raw image and a $\times 1$ whitened image show these cases, and both curves get close to 400 stored cost-to-goes. However, input given by a whitened image generally takes an order of magnitude fewer least-squares iterations to store the same number of cost-to-goes as a raw image, making it a computationally efficient image representation. The curve for a $\times 4$ raw image shows results for a natural image patch that has been upscaled by a factor of four using bicubic interpolation via Matlab's \texttt{imresize} function. The curve for a $\times 4$ sparse code shows results for an overcomplete sparse code. Each of these representations has 1600 pixels (i.e., $4\times 400$ pixels, treating the coefficients of a Gabor function basis in the same way as double precision pixels in an image). Increasing the number of weights to 1600 allows a linear network to store a maximum of 1600 cost-to-go values. However, the best result for the $\times 4$ resized image remains below 400 stored cost-to-goes, and uses approximately the same number of least squares iterations as the best result for the $\times 4$ sparse code; which gets close to 1600 stored cost-to-goes (the best result in Figure~\ref{result1} is 1323 cost-to-goes for 2701 least squares iterations). 

These results confirm that computationally efficient image representations must be both overcomplete \emph{and} sparse to increase the storage capacity of a linear network. The key role of sparsity is discussed in \cite{loxley2}, where it is shown how a sparse code decorrelates neighbouring pixels in an image and leads to an increase in the column-space dimension (rank) of the design matrix for the associated least squares problem. A sparse code also leads to a well-conditioned Hessian matrix, reducing the number of least squares iterations required for convergence. Since standard decorrelation techniques do not generalize to overcomplete representations, an overcomplete sparse code seems to be a minimum requirement for increasing neural network storage capacity efficiently.

Surprisingly, if we continue to increase the overcompleteness of image representations, the network storage capacity continues to increase. This is shown in Figure~\ref{result2} for network inputs given by $\times 4$, $\times 16$, and $\times 64$ overcomplete sparse codes. The size of each image patch is 361 pixels; so the $\times 4$ sparse code has $4\times 361=1444$ pixels, the $\times 16$ sparse code has $16\times 361=5776$ pixels, and the $\times 64$ sparse code has $64\times 361=23104$ pixels. These quantities are shown as dashed vertical lines in Figure~\ref{result2}. The general result in Figure~\ref{result2} is that the number of stored cost-to-goes (i.e., the network storage capacity) increases as the overcompleteness of an image representation increases. The first vertical dashed line at 361 cost-to-go values corresponds to the maximum storage capacity of a linear network with inputs given by complete codes. The other three vertical dashed lines correspond to maximum storage capacities for inputs given by $\times 4,\ \times 16,$ and $\times 64$ overcomplete sparse codes, respectively. 

The best result for each representation in Figure~\ref{result2} is summarized in Table~\ref{t1}. An extra data point (not shown in Figure~\ref{result2}) shows the limit of our calculation at 21675 stored cost-to-go values. This limit holds because we have a maximum of 22201 image patches, while the parameterization of the benchmark set $D$ implemented here means the next data point is a benchmark with 22707 states -- more than the available number of image patches. 
\begin{table}
\centering
\begin{tabular}{l|r|r}
Representation&$\#$ LS Iterations& $\#$ Stored Cost-To-Goes\\
\hline
\hline
$\times 1$ whitened image& 177 &243 \\
$\times 4$ sparse code& 2963&1323\\
$\times 16$ sparse code& 4527&5043\\
$\times 64$ sparse code& 9679&19683\\
\hline
\hline
$\times 64$ sparse code& 20009&21675\\
\end{tabular}
\\\vspace{10pt}
\caption{(Top) Four data points in Figure~\ref{result2}. (Bottom) One data point not shown in Figure ~\ref{result2}.}\label{t1}
\end{table}
Some variation in the average number of least-squares iterations can be seen in Table~\ref{t1}, particularly for the $\times 4$ sparse code. This is due to the relative proximity of each data point to its limiting case given by each of the vertical dashed lines. For example, the best result for the $\times 4$ sparse code is the data point at 1323 stored cost-to-go values, which is closer to its limiting value of 1444 than any other image representation is to its limiting value. At this point the condition number of the Hessian matrix is relatively large and, as a result, a larger number of least squares iterations is required.

\subsection{Optimal and Greedy Policies in the Benchmark}\label{horizon}
\begin{figure}[] 
\centering
\includegraphics[scale=0.6,clip=true,bb=50 280 500 600]{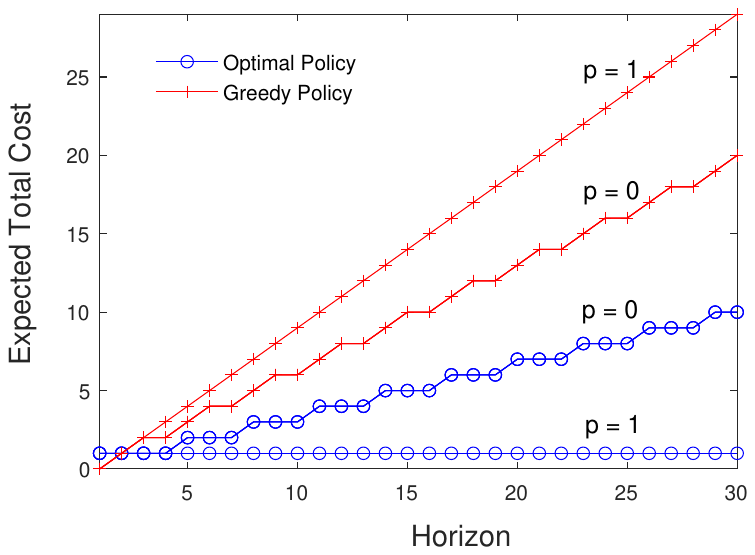}
\caption{Expected total cost of the optimal policy from Table \ref{t2} (blue circles), and the greedy policy from Table \ref{t3} (red pluses), as the horizon goes from $N=1$ to 30 time periods for $p=0$ and $p=1$.}
\label{result5}
\end{figure}

Evasive target dynamics in the benchmark target-tracking task leads to different policies for optimal and greedy tracking. The Markov chain in Appendix A1 depends on a single parameter $p$. When $p=0$, the target dynamics is deterministic and periodic and given by the repeating sequence $(\mathrm{sdr})^*$; where $\mathrm{s}\rightarrow (0,0)$, $\mathrm{d}\rightarrow (1,1)$, and $\mathrm{r}\rightarrow (0,1)$. The optimal policy from fitted value iteration is given in Table~\ref{t2} when the chain is initially at $\mathrm{s}$, and cycles between three states: $i_k=((0,1),(0,0))$, $i_{k+1}=((0,0),(1,1))$, and $i_{k+2}=((0,0),(0,1))$. For each state $i=(a_1,b_1)$, the first pair $a_1$ is the coordinate difference between the target and tracker (controller), while the second pair $b_1$ is the target's change in coordinates during the previous time period. 
\begin{table}
\centering
\begin{tabular}{c|c|c}
State&Control&Cost \\
\hline
\hline
$((0,1),(0,0))$&$(1,0)$ &1 \\
$((0,0),(1,1))$&$(0,1)$ &0 \\
$((0,0),(0,1))$&$(0,1)$ &0 
\end{tabular}
\\\vspace{10pt}
\caption{Optimal policy $\mu^*$ and cost per time period $g$.}\label{t2}
\end{table}
Only the first pair $a_1$ can have a non-zero cost, given by $\lVert a_1\rVert_2^2$ in (\ref{bmt}), and therefore the total cost of the optimal policy increases by one every three time periods. This is shown in Figure~\ref{result5} for $p=0$. 

The greedy policy can be found by replacing $\bar{J}_{k+1}$ in (\ref{bmt}) with $\bar{g}_{k+1}$, leading to the expression:
\begin{equation}
\underset{u^{G}\in U}{\operatorname{min}}\ \sum_{b_2} p(b_2|b_1)\lVert a_1+u^{G}-b_2\rVert_2^2.\label{grd}
\end{equation}
According to (\ref{grd}), a greedy tracker chooses $u^G$ as close as possible to $b_2-a_1$ (so that $c_{k+1}$ is as close as possible to $t_{k+1}$ at the next time period), without regard for what may happen at future time periods. That is, a greedy tracker has no scope for planning ahead. The greedy policy from (\ref{grd}) is given in Table~\ref{t3}. 
\begin{table}
\centering
\begin{tabular}{c|c|c}
State&Control&Cost \\
\hline
\hline
$((0,0),(0,0))$&$(1,0)$ &0 \\
$((0,-1),(1,1))$&$(0,1)$ &1 \\
$((0,-1),(0,1))$&$(0,1)$ &1 
\end{tabular}
\\\vspace{10pt}
\caption{Greedy policy $\mu^G$ and cost per time period $g$.}\label{t3}
\end{table}
The greedy policy also cycles between three states, with a total cost that increases by two every three time periods, as shown in Figure~\ref{result5} for $p=0$. This means the greedy policy has twice the total cost of the optimal policy when $p=0$, over any number of complete cycles.

When $p=1$, the target dynamics is given by the repeating sequence $\mathrm{r}^*$. The optimal policy in Table \ref{t2} is now $\mu^*((0,0),(0,1))=(0,1)$, with cost per time period $g((0,0),(0,1))=0$. Similarly, the greedy policy in Table \ref{t3} is now $\mu^G((0,-1),(0,1))=(0,1)$, with cost per time period $g((0,-1),(0,1))=1$. The ratio of total costs of the greedy to optimal policies is therefore $N-1$ to 1, so the greedy policy always has a cost $N-1$ times larger than that of the optimal policy, as seen in Figure~\ref{result5} for $p=1$.

When $0<p<1$, the target dynamics is stochastic, recurrent, and non-periodic: at the next time period, state $\mathrm{r}$ has probability $p$ of staying in the same state, and probability $1-p$ of going to state $\mathrm{s}$. The expected total cost of the optimal policy in Table \ref{t2} now lies between the curves (not shown) for the optimal policies with $p=1$ and $p=0$ in Figure~\ref{result5}. Similarly, the expected total cost of the greedy policy in Table \ref{t3} now lies between the curves (not shown) for the greedy policies with $p=0$ and $p=1$ in Figure~\ref{result5}. 

The optimal policy expected total cost in Figure \ref{result5} was found using both fitted value iteration and the dynamic programming algorithm, with both approaches yielding identical results. Proposition 1 allows us to freely switch between these two algorithms within the benchmark. The fact that the dynamic programming algorithm is an exact algorithm allows us to check the optimality of fitted value iteration for a particular image representation. This is advantageous when comparing different image representations for optimal control. Results for the greedy policy were found using policy evaluation; that is, by solving $J_{k}(i)=\sum_j p_{ij}(\mu_k^G(i))( g_k(i,\mu_k^G(i),j)+J_{k+1}(j))$ using dynamic programming iterations (see code at \cite{Loxley4}).

\subsection{Benchmark Performance with Length of Horizon}
The optimal and greedy policies given in Tables \ref{t2} and \ref{t3} turn out to be stationary policies for all values of $p$. This is confirmed using an infinite horizon analysis in Appendix A3. The cost of each policy is found by introducing a discount factor and applying policy evaluation in the limit of an infinite horizon (i.e., $N\rightarrow \infty$). The cost ratio of the greedy to optimal policies is completely consistent with our discussion of these policies for finite horizons. For example, the cost ratio becomes 2 when $p=0$, 5 when $p=0.75$, and diverges when $p=1$. For finite horizons, the first result ($p=0$) is most easily seen in Figure~\ref{result5} when $N=30$; giving a ratio of $20/10=2$. The final result ($p=1$) is also seen in Figure~\ref{result5} when $N=30$, giving a ratio of $29/1=29$. As $N$ increases, the numerator will continue to increase, diverging in the limit as $N\rightarrow\infty$. The result for $p=0.75$ can be seen from the ratio of expected total costs of the greedy and optimal results in Figure~\ref{result3}. 

\begin{figure}[] 
\centering
\includegraphics[scale=0.6,clip=true,bb=50 300 500 600]{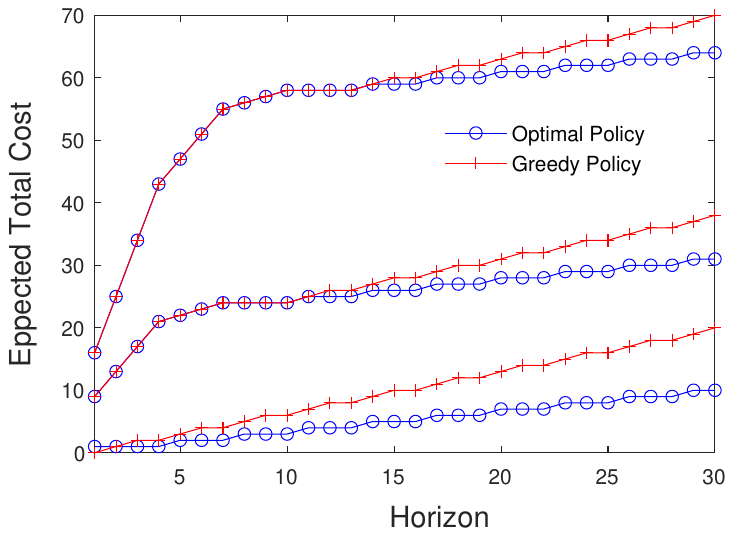}
\caption{Expected total cost of optimal and greedy policies as the horizon goes from $N=1$ to $N=30$ time periods for $p=0$ with different initial states.  The bottom pair of curves start at state $((0,1),(0,0))$; the middle pair start at state $((3,0),(0,0))$; and the top pair start at state $((4,0),(0,0))$.}
\label{result6}
\end{figure}

\begin{figure}[] 
\centering
\includegraphics[scale=0.65,clip=true,bb=50 280 500 600]{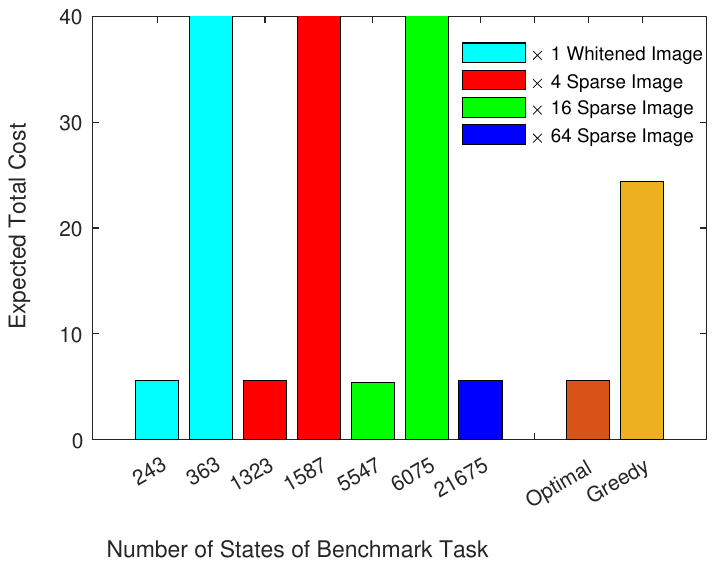}
\caption{Expected total cost of tracking a target in the benchmark over 30 time periods with $p=0.75$, versus number of states of the benchmark for different image-patch representations (see legend). Exact values for optimal and greedy target tracking (right columns) are shown for comparison.}
\label{result3}
\end{figure}

The optimal and greedy policies lead to more complicated dynamics over shorter horizons when different initial states are chosen. For an initial state given by $((0,1),(0,0))$ (i.e., where the tracker starts close to the target, as implied by the first pair of coordinates) the bottom pair of curves in Figure~\ref{result6} reach the stationary optimal and greedy policies in Tables \ref{t2} and \ref{t3} at very short horizons. As the initial distance between the tracker and target increases for positive coordinates (middle and top pairs of curves in Figure~\ref{result6}), the optimal and greedy policies take longer and longer to reach these stationary policies. For example, the top pair of curves has not reached their stationary policies until the horizon is around 14 time periods. In fact, it is found the dynamics of all initial states with one positive coordinate and one non-negative coordinate will eventually reach the stationary policies given in Tables \ref{t2} and \ref{t3} provided the horizon is long enough.

The greedy policy in Figure~\ref{result6} is shown to be an optimal policy when the initial state of the tracker is far from the target. Reaching a stationary policy only happens when the tracker gets close enough to the target (i.e., within a distance of one). Once the tracker is close to the target the greedy policy is no longer optimal (as shown at longer horizons in Figure~\ref{result6}). For initial states where both coordinates are negative, the tracker starts behind the target and can never catch up. It is then sufficient to apply a simple greedy approach to follow the target, ensuring the greedy policy is an optimal policy in this case. A more detailed discussion of initial-state dependence is given in Sec.~\ref{IS1}.

Now we are in a position to look at how the benchmark performs as the number of states increases. Increasing the number of states increases the size of the region over which a target may be tracked in the benchmark, and increases the number of stored cost-to-go values required for an optimal solution. From Figure~\ref{result2}, we would expect the use of efficient image representations should increase the number of states over which optimal target tracking is possible. Figure~\ref{result3} shows this result: the expected total cost of tracking a target with $p=0.75$ over thirty time periods is shown for various image representations as the number of states of the benchmark is increased. The initial states (taken from Tables \ref{t2} and \ref{t3}) are $((0,1),(0,0))$ for the optimal tracker, and $((0,0),(0,0))$ for the greedy tracker. For comparison, exact results are shown for both optimal and greedy target tracking in the two right columns in Figure~\ref{result3} -- as previously mentioned, for the benchmark we can compare fitted value iteration with the exact dynamic programming algorithm.

According to Figure~\ref{result3}, a representation given by a $\times 1$ whitened image can solve the benchmark optimally for 243 states, but not for 363 states -- the expected total cost now exceeds that for greedy target tracking. The reason is, as in Figure~\ref{result2}, a complete code stores a maximum of 361 cost-to-go values which is less than 363. Similarly, a $\times 4$ sparse code can solve the benchmark optimally for 1323 states, but not for 1587 ($>1444$) states; a $\times 16$ sparse code can solve the benchmark optimally for 5547 states, but not for 6075 ($>5776$) states; and a $\times 64$ sparse code can solve the benchmark optimally for 21675 states. This was the largest number of image patches tested in this work. Actual values for the number of states shown in Figure~\ref{result3} are due to the specific parameterization of the set $D$ implemented here.

\subsection{Benchmark Performance for Different Initial States}\label{IS1}
We now look in more detail at the effect of starting from different initial states in the reinforcement learning benchmark. This is done by starting the benchmark in an initial state, and then running it with either an optimal or greedy policy until reaching the horizon. The total cost of each policy can then be calculated and compared. This is repeated for each state of the benchmark.

\begin{figure}[] 
\centering
\includegraphics[scale=0.7,clip=true, bb=50 270 500 600]{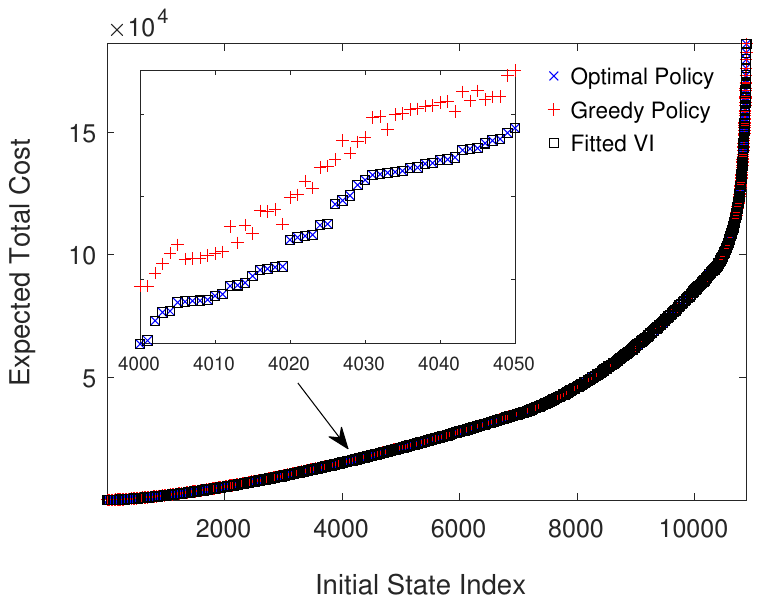}
\caption{Expected total cost of optimal and greedy policies for each initial state leading to a suboptimal greedy policy. These initial states are indexed from 1 to 10880. ``Fitted VI" is the policy found using fitted value iteration. Inset shows a close-up of initial states indexed from 4000 to 4050. The benchmark horizon is 200 time periods, and $p=0.4$ (see text for details).}
\label{result7}
\end{figure}

Training the benchmark with a $\times 64$ sparse code leads to 21675 possible initial states. The expected total cost of optimal and greedy policies is shown in Figure \ref{result7} for each initial state leading to a suboptimal solution under the greedy policy. These initial states are indexed from 1 to 10880 and comprise roughly half of the 21675 available initial states. For the remaining 10795 initial states (not shown) the greedy policy is also the optimal policy. The benchmark horizon is 200 time periods, and the Markov chain parameter is $p=0.4$. Optimal policies are found using both fitted value iteration and exact dynamic programming, as shown in Figure \ref{result7} (Inset) for initial states indexed from 4000 to 4050.

Looking at a few specific initial states helps to understand Figure \ref{result7}. The lowest cost optimal policy in  Figure \ref{result7} starts at state $((0,0),(1,1))$ and has an expected total cost of 54. We can derive a theoretical bound for this state since it is part of the stationary policy in Table \ref{t2} (see also Figure \ref{result5}). For $p=0$, its total cost is given by $(200-2)/3\times 1=66$. From Figure \ref{result5}, we know the optimal cost for $p=0.4$ would be less than that for $p=0$, leading to $54<66$. We can also predict a value for the lowest cost greedy policy by making use of the infinite horizon result for the total cost ratio from Appendix A3: $(1+1/(1-p))$. The expected total cost of the greedy policy can then be found by multiplying this ratio by the expected total cost of the optimal policy: for $p=0.4$ we get $(1+1/(1-p))\times 54=144$. This exactly matches the value in Figure \ref{result7} (which cannot be seen due to the plot scale).

Now that we have some theoretical justification for the lowest cost policy in Figure \ref{result7}, let us attempt to understand the highest cost policy in that figure. This policy starts at state $((42, -42),(1,1))$ and has an expected total cost of $185870$ for the optimal policy, and $185890$ for the greedy policy. We can understand this as follows. Each number in the coordinate pair for $c_k-t_k\in D$ has the range $[-42,42]$ when the number of states is 21675. Starting at state $((42, -42),(1,1))$ means the tracker is as far behind the target as possible, but it is also as far to the right of the target as possible. Consider the corresponding situation when $p=0$: during three time periods, the target will move right once and up twice. If the tracker simply moves up three times over the same three time periods, then it has reduced the distance to the target by one every three time periods. Therefore, it can reach the target within $42\times 3=126$ time periods, and apply the stationary optimal policy thereafter. There is no other initial state with a suboptimal greedy policy that can have a larger expected total cost over this horizon. The same argument holds approximately when $p=0.4$. 

For the other 10795 initial states where the greedy policy is the optimal policy (not shown), the lowest cost policy starts at state $((0,0),(0,0))$ and has an expected total cost of 145. This is close to the expected total cost of the greedy policy for the initial state $((0,0),(1,1))$ just discussed. Predictably, the highest cost policy starts with the tracker as far behind the target as possible, in state $((-42,-42),(1,1))$, and has an expected total cost of $701100$. 

\subsection{Benchmark Performance with Partial Training Sets}\label{IS2}
Now we consider situations where only some of the states (image patches) of the benchmark are used to train the network. One possibility would be to focus training on initial states that have a suboptimal greedy policy. However, under an optimal policy states outside this set will be visited. A better approach makes use of the analysis in the previous section to partition the benchmark states into two sets. States with at least one non-negative coordinate for $c_k-t_k$ in the benchmark correspond to a situation where the tracker is not both behind and to the left of the target. These states are independent of the set of states where the tracker is both behind and to the left of the target: in which case the tracker never reaches the target. When states partition in this way, it is possible to focus training on the relevant partition to achieve optimal performance. 
\begin{figure}[] 
\centering
\includegraphics[scale=0.7,clip=true, bb=50 270 500 600]{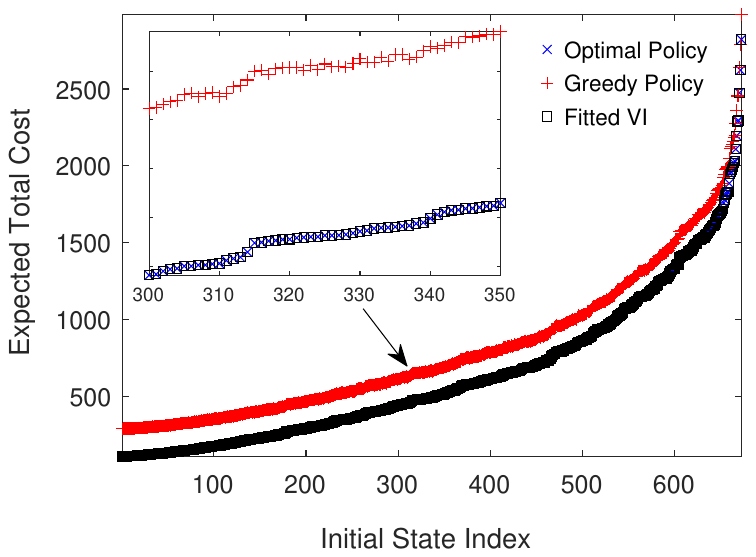}
\caption{Expected total cost of optimal and greedy policies for each initial state with a suboptimal greedy policy when only 78\% of the training set is used (see Text for details). Initial states are indexed from 1 to 672. Fitted value iteration (Fitted VI) remains optimal (Inset). The benchmark horizon is 400 time periods, and $p=0.4$.}
\label{result9}
\end{figure}

In Figure \ref{result9}, the benchmark is used with a $\times 4$ sparse code leading to 1323 states (a smaller number of states is used here as it is easier to display). Of these, 1031 states have at least one non-negative coordinate (i.e., 78\% of the total number  of states). These states are used as the training set. Fitted value iteration is then applied to find optimal policies for the subset of 672 initial states with suboptimal greedy policies. Figure \ref{result9} shows these policies remain optimal under the proposed training strategy. However, further decreasing the size of the training set rapidly degrades performance as states from the relevant partition start to be left out of the training set. The benchmark provides an ideal testing ground to investigate these types of behaviours.

\section{Conclusion and Discussion}\label{conc}

Optimal control over a sequence of natural images provides some new and interesting interpretations of the general reinforcement learning methodology. 

The first consideration is what state information to include for optimal control over natural image sequences. The most obvious condition is that an image must contain necessary task-relevant information leading to the assignment of a cost or reward at each time period. However, this is not the only condition. We also saw that any model of the environment dynamics must be capable of generating images from that environment. This is essential because a controllable Markov chain, or a Markov decision process, must provide a process for generating the next state from the current state and a chosen control. In the present case, this process gives the dynamics of moving through an environment described by natural images.

The second consideration is to ensure that reinforcement learning can be done efficiently with natural images. The key is to find an image representation that allows reinforcement learning algorithms to become computationally efficient. Here, we made use of the idea of efficient neural representations of visual stimuli known as sparse codes. To test this idea required developing a new reinforcement learning benchmark to compare different image representations for computational efficiency during optimal control. This new benchmark had to be simple to implement, to easily scale to large numbers of states and long horizons, and to include the coexistence of optimal and suboptimal polices for the purpose of demonstrating optimality. These properties were necessary to convincingly demonstrate computational efficiency for large tasks with many states.

The main contribution of this work was to convincingly demonstrate the computational efficiency of representing natural images as overcomplete sparse codes in reinforcement learning tasks. This is a new and significant result. The best result presented here was finding an optimal policy using a $\times 64$ overcomplete sparse code, allowing for the solution of an optimal control task approximately three orders of magnitude larger (in base 4) than could be solved using a complete code such as one generated by ICA or PCA. Theoretical justification relates to the properties of the Hessian and design matrices of the corresponding linear least squares problem. Provided the number of images is no larger than the size of an overcomplete sparse code, the design matrix is approximately square. A sparse code approximately decorrelates neighbouring image pixels, thereby conditioning the Hessian matrix and improving the rank of the design matrix. The effect is to maximize the storage capacity of a linear network, and reduce the number of least squares iterations required for training. Decorrelation techniques do not usually generalize to overcomplete representations, so sparsity and overcompleteness seem to be minimum requirements to achieve this effect. An interesting question then becomes: what is the limit to overcompleteness for a sparse code?

This work also shows deep learning is not necessary for efficient optimal control with natural images. A deep neural network may be able to learn specific features for a specific image dataset to help with optimal control. However, a sparse code only depends on the statistics of natural images, and therefore provides a good set of general purpose features that can be used on any image dataset. Of course, deep learning can also be used to learn sparse codes \citep{papyan,li}. Our model has the advantage that it is simpler to train and uses less memory than a typical deep neural network. It should be noted that sparse codes can also be generated directly from hardware using purpose built low-power neuromorphic computer chips \citep{chip}.

The time complexity of training our model depends upon the solution of two linear least squares problems. The first least squares problem has a $d\times m$ design matrix (for a single image with $d$ pixels and a sparse code of size $m$), leading to $O(dmt_1)$ for an iterative algorithm such as gradient descent; with $t_1$ the number of iterations until convergence. For $n$ images this must be repeated $n$ times, leading to $O(dmt_1n)$. The second least squares problem has a $m\times n$ design matrix, leading to $O(mnt_2)$; with $t_2$ the number of iterations until convergence. For optimal control over $N$ time periods this must be repeated $N$ times, leading to $O(mnt_2N)$. The overall time complexity of training is therefore $O(mn(dt_1 + Nt_2))$. Both least squares problems are well-conditioned when Gabor functions and sparse codes are used (at least until $n\rightarrow m$, see Figure \ref{result2}) so the iteration numbers $t_1$ and $t_2$ are not excessive. Our model only requires memory to store $m(d+1)$ weights. 

The original motivation for this work was to investigate possible computational advantages of the efficient coding hypothesis from neuroscience \citep{barlow}. If neural codes do in fact make use of sparseness and overcompleteness, this work has demonstrated some computational advantages to doing so. In particular, the decorrelating effect of a sparse code increases the speed of learning, while an overcomplete sparse code can additionally take advantage of an increase in the number of network weights to efficiently increase its memory capacity. These findings suggest two possible computational advantages of sparse codes in neuroscience.

\subsection*{Acknowledgements}
I would like to thank J.~Daugman for his interest and encouragement in this project. I would also like to thank B.~Olshausen for graciously hosting several sabbatical visits over this time.

\bibliographystyle{plainnat}
\bibliography{Loxley_references} 

\newpage

\section*{Appendix A1: Benchmark Target Tracking Task}

\subsection*{Target Dynamics}

The target dynamics for the benchmark has two key requirements: 1) the dynamics must easily scale to long horizons and many states, and 2) the target must be difficult enough to track so that simple algorithms such as greedy search perform suboptimally. Both points can be addressed by introducing a simple generative model for the target dynamics.  

Motivated by the dragonfly video dataset used for optimal control in \citet{loxley2}, the following type of scenario is proposed. The target (T) starts on the bottom row of a $3\times 3$ grid (shown in Figure \ref{fig1}, first grid). The target and the tracker can both move up, left, or right at each time period. However, the target (but not the tracker) can also make an ``evade" move by moving diagonally to evade the tracker. The tracker must therefore carefully plan its sequence of moves in order to follow the target closely. 

During the first time period (second grid in Figure \ref{fig1}) the target remains where it is, while at the beginning of the second time period (third grid in Figure \ref{fig1}) it makes a diagonal move. An optimal tracker (blue square) anticipates a diagonal move immediately following a time period where the target is stationary, and pays a small cost to move up one square away from the target during the first time period (second grid in Figure \ref{fig1}). Whether the target next moves diagonally left or diagonally right, the optimal tracker is now able to reach the target during the next time period (third grid in Figure \ref{fig1}). On the other hand, a tracker applying a greedy algorithm (red squares) tries to stay as close to the target as possible during each time period, allowing the target to ``evade" the greedy tracker when it makes its diagonal move, causing the greedy tracker to fall one move behind the target (third and fourth grids in Figure \ref{fig1}). A greedy tracker will always be suboptimal for this sequence.
\begin{figure}
\center
\begin{tikzpicture} [scale = 1.5]
\draw[step=0.5cm,color=gray] (-0.5,-0.5) grid (1,1);
\node[fill=blue!50, minimum size=7.5mm] at (0.25,-0.25) {};
\node[fill=red, minimum size=5mm] at (0.25,-0.25) {\large T};
\end{tikzpicture}\hspace{30pt}
\begin{tikzpicture} [scale = 1.5]
\draw[step=0.5cm,color=gray] (-0.5,-0.5) grid (1,1);
\node[fill=red, minimum size=7.5mm] at (0.25,-0.25) {\large T};
\node[fill=blue!50, minimum size=7.5mm] at (0.25,0.25) {};
\end{tikzpicture}\hspace{30pt}
\begin{tikzpicture} [scale = 1.5]
\draw[step=0.5cm,color=gray] (-0.5,-0.5) grid (1,1);
\node[fill=red, minimum size=7.5mm] at (0.25,0.25) {};
\node[fill=blue!50, minimum size=7.5mm] at (0.75,0.25) {\large T};
\end{tikzpicture}\hspace{30pt}
\begin{tikzpicture} [scale = 1.5]
\draw[step=0.5cm,color=gray] (-0.5,-0.5) grid (1,1);
\node[fill=red, minimum size=7.5mm] at (0.75,0.25) {};
\node[fill=blue!50, minimum size=7.5mm] at (0.75,0.75) {\large T};
\end{tikzpicture}\hspace{30pt}
\caption{A target tracking sequence (from left to right) showing optimal and suboptimal trackers. A tracker can move either ``up" or ``right", while the target can move ``up", ``right", or ``diagonally". The suboptimal tracker (red square) follows the target (T) as closely as possible at each step (a greedy approach), causing it to fall behind when the target moves diagonally. The optimal tracker (blue square) follows the target by anticipating a diagonal move following a time period where the target is stationary.}
\label{fig1}
\end{figure}

The sequence of target moves in Figure \ref{fig1} can be generalized and modelled as a regular language \citep{sipser}. The alphabet of target-moves is $\{\mathrm{s,d,r}\}$; where $\mathrm{s}\rightarrow (0,0)$, $\mathrm{d}\rightarrow (1,1)\ \text{or}\ (-1,1)$, and $\mathrm{r}\rightarrow (0,1)\ \text{or} \ (-1,0)\ \text{or}\ (1,0)$. Each symbol in the alphabet represents a change in the target coordinates, $\Delta t$, as the target moves from its current position to its next position. The regular expression given by \mbox{$\mathrm{r}^*\cup (\mathrm{sdr})^*$} generates all strings in this language, representing all possible target dynamics. For example, the string given by ``$\mathrm{sdr}$" represents the sequence in Figure \ref{fig1}: $\Delta t_1=(0,0)$; $\Delta t_2=(1,1)$; and $\Delta t_3=(0,1)$. According to this sequence, a target starting at $(2,1)$ stays at $(2,1)$ during the first time period; then moves to $(3,2),$ and $(3,3)$ in the second, and third time periods, respectively (see Figure \ref{fig1}).

A Markov chain can be constructed to generate stochastic dynamics obeying this language. The Markov chain is parameterized by the single degree of freedom available in the regular expression: whether a symbol of a substring begins with an ``$\mathrm{r}$" or an ``$\mathrm{s}$". This is parameterized using the probability $p$, leading to the Makov chain shown in Figure \ref{fig2}. 
\begin{figure}
\centering
\begin{tikzpicture}[->,shorten >=2pt,line width=0.5pt,node distance=2cm]
\node [circle,draw] (r) at (0.0,0.3) {r};
\path (r) edge [loop above] node [above] {$p$} (rloop);
\node [circle,draw] (d) at (-0.8,-1) {d};
\path (d) edge [bend left] node [above left] {$1$} (r);
\node [circle,draw] (s) at (0.8,-1) {s};
\path (r) edge [bend left] node [above right] {$1-p$} (s);
\path (s) edge [bend left] node [below] {$1$} (d);
\end{tikzpicture}
\caption{A Markov chain for generating the target dynamics shown in Figure \ref{fig1}. In Figure \ref{fig1}, the states $\mathrm{s,d,}$ and $\mathrm{r}$ correspond to ``same position", ``diagonal move", and ``up" or ``right", respectively. The non-zero transition probabilities are: \mbox{$p(\Delta t_k=\mathrm{d}|\Delta t_{k-1}=\mathrm{s})=1$}, \mbox{$p(\Delta t_k=\mathrm{r}|\Delta t_{k-1}=\mathrm{d})=1$}, \mbox{$p(\Delta t_k=\mathrm{s}|\Delta t_{k-1}=\mathrm{r})=1-p$}, and \mbox{$p(\Delta t_k=\mathrm{r}|\Delta t_{k-1}=\mathrm{r})=p$}.}
\label{fig2}
\end{figure}
Different values of $p$ give different steady state distributions of the Markov chain. When $p=0$ or 1, the target dynamics is a deterministic periodic sequence: $(\mathrm{sdr})^*$ for $p=0$, and $\mathrm{r}^*$ for $p=1$. When $0<p<1$, the Markov chain is recurrent and non-periodic: i.e., when $p=0.5$ the steady state probabilities of $\mathrm{s,d,}$ and $\mathrm{r}$ are $1/4$, $1/4$, and $1/2$, respectively; and a typical sequence of twelve target moves looks like $\mathrm{sdrrsdrrrrsdr}$.

\subsection*{Target Tracking}
It is now possible to specify the target tracking task used in the reinforcement benchmark. This task requires two dynamical variables $t_k,c_k\in\mathbb{Z}^2$ representing the discrete two-dimensional coordinates of a target, and a controller (tracker), respectively, at time period $k$. Changes to the environment are described by the discrete dynamical system:
\begin{align}
&t_{k+1}=t_k+\Delta t_k,\label{sys1}\\
&c_{k+1}=c_k+u_k.\label{sys2}
\end{align}
A target updates its position from $t_k$ to $t_{k+1}$ according to the transition probabilities of the Markov chain presented in the previous section. In order to follow the target as closely as possible a controller updates its position from $c_k$ to $c_{k+1}$ by choosing a control $u_k$ from the set $U$ of available controls. The state information required for choosing an optimal control is given by defining the state to be:
\begin{equation}
i_k=(c_k-t_k,\Delta t_{k-1})\in D \times T,\label{state1}
\end{equation}
where $c_k-t_k\in D$, and $\Delta t_{k-1}\in T$; so that $S=D \times T$. Here, $D,T,U\subset \mathbb{Z}^2$ are sets of integer pairs. 
In order to derive the transition probabilities $p_{ij}(u)$, we proceed as follows. According to (\ref{state1}), and making use of (\ref{sys1}) and (\ref{sys2}), the state update is given by
\begin{align}
i_{k+1}&= (c_{k+1}-t_{k+1},\Delta t_k),\nonumber\\
&=(c_k-t_k+u_k-\Delta t_k,\Delta t_k).\label{state2}
\end{align}
Denoting the state $i=i_k$ and its successor $j=i_{k+1}$ in terms of their individual components:
\begin{equation}
i=(a_1,b_1),\ \ \ \ j=(a_2,b_2),\label{state3}
\end{equation}
and upon equating with (\ref{state1}) and (\ref{state2}), leads to $a_1=c_k-t_k,b_1=\Delta t_{k-1},a_2=a_1+u_k-\Delta t_k,$ and $b_2=\Delta t_k$. These equations can be viewed as constraints on the allowed values of $j$ via $a_2$ and $b_2$, and represented as \mbox{$\delta(a_2 - (a_1+u_k-\Delta t_k))$} and \mbox{$\delta(b_2 - \Delta t_k)$}; where $\delta(k)=1$ if $k=0$, and $\delta(k)=0$ otherwise. The controllable Markov chain is governed by the the Markov chain from Figure~\ref{fig2} describing the target dynamics: $p(\Delta t_k|\Delta t_{k-1})$, as well as the constraints on $j$, giving:
\begin{align}
p_{ij}(u)&=\sum_{\Delta t_k}p(\Delta t_k|b_1)\delta(a_2 - (a_1+u-\Delta t_k))\delta(b_2 - \Delta t_k),\\
&=p(b_2|b_1)\delta(a_2 - (a_1+u-b_2)).\label{trackdyn}
\end{align} 
The cost per time period is the distance between the tracker and the target at each time period, and is given by 
\begin{equation}
g_k(i,u,j)=\lVert a_1\rVert_2^2,\label{trackcost}
\end{equation}
where $\lVert.\rVert_2$ is the Euclidean distance. For the benchmark tracking task, the sets $T$ and $U$ are specified by $T=\{(0,0),(1,1),(0,1)\}$ and $U=\{(0,0),(1,0),(0,1)\}$. It is necessary to restrict the target moves as in $T$, otherwise too much uncertainty builds up in the target dynamics to achieve good optimal policies for the tracking task. The set $D$ is the set of all coordinate differences (i.e., distances) between the target and the controller. The size of this set is entirely flexible, and can either be increased or decreased according to the number of states required for the benchmark.

\section*{Appendix A2: Method for generating overcomplete sparse codes}\label{gabmeth}
The method used here to generate overcomplete sparse codes of a natural images is taken from \cite{loxley1,loxley2}, and starts with the (real-valued) two-dimensional (2D) Gabor function given by: 
\begin{equation}
G({r},{r}^{\prime})=A\exp{\left[-\frac{1}{2}\left(\frac{\tilde{i}^2}{\sigma_{x}({r}^{\prime})^2}+\frac{\tilde{j}^2}{\sigma_{y}({r}^{\prime})^2}\right)\right]}\cos{\left[k({r}^{\prime})\tilde{j}+\varphi({r}^{\prime})\right]},\label{dic1}
\end{equation}
and
\begin{equation}
(\tilde{i},\tilde{j})=\left(\begin{array}{cc}
\cos{\phi({r}^{\prime})}&-\sin{\phi({r}^{\prime})}\\
\sin{\phi({r}^{\prime})}&\cos{\phi({r}^{\prime})}\\
\end{array}
\right)
\left(\begin{array}{c}
i-x_0({r}^{\prime})\\
j-y_0({r}^{\prime})\\
\end{array}
\right),\label{dic2}
\end{equation}\\
where $k({r}^{\prime}) = 2\pi/\lambda({r}^{\prime})$; and where ${r}=(i,j)$ and ${r}^{\prime}=(i',j')$ are discrete two-dimensional coordinates. When the 2D Gabor function is adapted to natural image statistics, the three spatial Gabor parameters $\sigma_x,\sigma_y,$ and $\lambda$ are found to be strongly correlated and have heavy-tailed distributions \citep{loxley1}. The joint probability density of these parameter values is approximated using the sampling scheme in Table \ref{cop}, and described by a Gaussian copula with Pareto marginal distributions.
\begin{table}
\centering
\begin{tabular}{ll}
\hline
Gabor Parameter(s)& Sample Transformation\\
\hline
$\sigma_x^\prime,\sigma_y^\prime,\lambda^\prime$&$(\sigma_x^\prime,\sigma_y^\prime,\lambda^\prime)=(1,1,\rho)z$\\
$\sigma_x$& $\sigma_x={\cal{PCDF}}^{-1}({\cal{NCDF}}(\sigma_x^\prime|0,1)|\alpha_1,\beta_1)$\\
$\sigma_y$& $\sigma_y={\cal{PCDF}}^{-1}({\cal{NCDF}}(\sigma_y^\prime|0,1)|\alpha_2,\beta_2)$\\
$\lambda$& $\lambda={\cal{PCDF}}^{-1}({\cal{NCDF}}(\lambda^\prime|0,1)|\alpha_3,\beta_3)$\\
\hline
\end{tabular}
\vspace{10pt}
\caption{Sampling scheme for the three spatial Gabor function parameters: sample $z\sim {\cal{N}}(0,1)$ from the standard normal distribution, then apply the parameter transformations listed in the table. Here, $\rho, \alpha_i$, and $\beta_i$ are model parameters, ${\cal{PCDF}}^{-1}(x|\alpha,\beta)=\frac{\beta}{(1-x)^{1/\alpha}}$ is the inverse CDF for the Pareto distribution, and ${\cal{NCDF}}(x|0,1)$ denotes the CDF for the standard normal distribution.}\label{cop}
\end{table} 
Due to the Pareto marginal distributions the sampling scheme is length scale invariant. Scale invariance is a key property of natural images. The resulting set of randomly generated Gabor functions are therefore self-similar and multiscale in the same way as a self-similar multiresolution wavelet scheme. All other Gabor parameters are sampled uniformly over their respective ranges. 

In the first step, a sample is collected for each of the seven Gabor parameters $(\phi,\varphi,\sigma_{x},\sigma_{y},\lambda,x_0,y_0)$, leading to a single 2D Gabor function indexed by a value of ${r}^{\prime}$. This step is repeated $m$ times; leading to $m$ Gabor functions indexed by $m$ unique values of ${r}^{\prime}$. Two-dimensional Gabor functions are not orthogonal. However, given an image $I({r})\in\mathbb{R}^d$, it is possible to find its sparse code $a({r}^{\prime})\in \mathbb{R}^m$ using a least-squares approximation. Letting $G\in\mathbb{R}^{d\times m}$ be a matrix with elements $G({r},{r}^{\prime})$, the sparse code $a({r}^{\prime})$ is found by solving the least-squares problem,
\begin{align*}
&\text{minimize}\ \ \ \lVert Ga - I \rVert_2^{2}\\
&\text{subject to}\ \ \ a\in \mathbb{R}^m.
\end{align*}
When $m>d$, the sparse code is overcomplete; meaning that there are more Gabor function coefficients than image pixels.

\section*{Appendix A3: Infinite horizon policy evaluation}
A stationary policy is a policy that does not change with each time period $k$, and can therefore be written as $\pi=\{\mu(i),\mu(i),...\}$. In Sec.~\ref{horizon}, we identified two stationary policies. We would now like to find the their expected total costs in the infinite horizon limit (i.e., as $N\rightarrow\infty$). To work in this limit, we must introduce a \emph{discount factor} $0<\alpha\leq 1$ that prevents total costs from becoming infinite. For any stationary policy $\mu$, the expected total costs $J_\mu(1),...,J_\mu(n)$ are then found as the unique solution to the following system of linear equations:
\begin{equation}
J_{\mu}(i)=g(i)+\alpha\sum_j p_{ij}(\mu(i))J_{\mu}(j).\label{policyeval}
\end{equation}
The stationary optimal policy identified in Sec.~\ref{horizon} is given by $\mu^*((0,1),(0,0))=(1,0)$, $\mu^*((0,0),(1,1))=(0,1)$, and $\mu^*((0,0),(0,1))=(0,1)$. These states have a cost per time period given by $g((0,1),(0,0))=1$, $g((0,0),(1,1))=0$, and $g((0,0),(0,1))=0$. After rearranging (\ref{policyeval}) as
\begin{equation}
\sum_j\big(\delta_{ij} - \alpha p_{ij}(\mu(i))\big)J_{\mu}(j)=g(i),\label{linearsyst}
\end{equation}
the linear system can be written as $\boldsymbol{AJ}_\mu=\boldsymbol{g}$, where 
\begin{equation}
\boldsymbol{A}=
\begin{bmatrix}
1&-\alpha&0\\
0&1&-\alpha\\
-\alpha(1-p)&0&1-\alpha p
\end{bmatrix},
\end{equation}
so that $\det{(\boldsymbol{A})}=1-\alpha p-\alpha^3(1-p)$. For the stationary optimal policy, $\boldsymbol{g}$ is given by $[1, 0, 0]^{\top}$, and it is easily confirmed that $\boldsymbol{AJ}_\mu^*=[1, 0, 0]^{\top}$ is solved by
\begin{equation}
\boldsymbol{J}_\mu^*=
\begin{bmatrix}
1-\alpha p\\
\alpha^2(1-p)\\
\alpha(1-p)
\end{bmatrix}
\times(1-\alpha p-\alpha^3(1-p))^{-1}.
\end{equation}

The stationary greedy policy identified in Sec.~\ref{horizon} is given by $\mu^G((0,0),(0,0))=(1,0)$, $\mu^G((0,-1),(1,1))=(0,1)$, and $\mu^G((0,-1),(0,1))=(0,1)$. These states have a cost per time period given by $g((0,0),(0,0))=0$, $g((0,-1),(1,1))=1$, and $g((0,-1),(0,1))=1$. Now $\boldsymbol{g}$ is given by $[0, 1, 1]^{\top}$. It can then be confirmed that $\boldsymbol{AJ}_\mu^G=[0, 1, 1]^{\top}$ is solved by 
\begin{equation}
\boldsymbol{J}_\mu^G=
\begin{bmatrix}
\alpha+\alpha^2(1-p)\\
1+\alpha(1-p)\\
1+\alpha^2(1-p)
\end{bmatrix}
\times(1-\alpha p-\alpha^3(1-p))^{-1}.
\end{equation}

The expected total cost ratio $\boldsymbol{J}_\mu^G/\boldsymbol{J}_\mu^*$, for the first greedy state $i^G_1=(0,0),(0,0)$, and the first optimal state $i^*_1=(0,1),(0,0)$, is
\begin{equation}
\frac{J_\mu^G(i^G_1)}{J_\mu^*(i^*_1)} = \alpha+\frac{\alpha^2}{1-\alpha p}.\label{orig}
\end{equation}
Taking the limit of zero discounting (i.e., $\alpha\rightarrow 1$), (\ref{orig}) becomes 
\begin{equation}
\lim_{\alpha\rightarrow 1}\frac{J_\mu^G(i^G_1)}{J_\mu^*(i^*_1)} = 1+\frac{1}{1-p}.
\end{equation}
This ratio describes the gap in the expected total cost between the greedy and optimal policies in the infinite horizon limit, and is useful for comparing with the finite horizon results of Sec~\ref{horizon}. Some useful values include: 2 (when $p=0$), 5 (when $p=0.75$), and $1/(1-p)$ as $p\rightarrow 1$. Alternatively, for $p=1$, the original expression (\ref{orig}) gives $1/(1-\alpha)$ as $\alpha\rightarrow 1$.

\end{document}